\def\year{2018}\relax
\documentclass[letterpaper]{article} 
\usepackage{aaai18}  
\usepackage{times}  
\usepackage{amsmath}
\usepackage{helvet}  
\usepackage{courier}  
\usepackage{amssymb}
\usepackage{url}  
\usepackage{graphicx}  
\usepackage{subcaption,mwe}
\usepackage{comment}
\usepackage{color}
\usepackage{multirow}
\usepackage{enumitem}

\newcommand{\eat}[1]{}
\newcommand{\nb}[1]{\textcolor{blue}{}}
\newcommand{\nc}[1]{\textcolor{red}{}}
\newcommand{\nw}[1]{\textcolor{green}{}}

\newcommand{\citet}[1]{\citeauthor{#1}~\shortcite{#1}}
\newcommand{\citep}{\cite}

\begin{document}
%
\title{Event Representations with Tensor-based Compositions}

\author{Noah Weber \\Stony Brook University \\Stony Brook, New York, USA\\nwweber@cs.stonybrook.edu \And Niranjan Balasubramanian \\Stony Brook University \\Stony Brook, New York, USA\\niranjan@cs.stonybrook.edu \And Nathanael Chambers
\\United States Naval Academy\\Annapolis, Maryland, USA\\nchamber@usna.edu
}
\maketitle
\begin{abstract}
Robust and flexible event representations are important to many core areas in language understanding. Scripts were proposed early on as a way of representing sequences of events for such understanding, and has recently attracted renewed attention. However, obtaining effective representations for modeling script-like event sequences is challenging. It requires representations that can capture event-level and scenario-level semantics. We propose a new tensor-based composition method for creating event representations. The method captures more subtle semantic interactions between an event and its entities and yields representations that are effective at multiple event-related tasks. With the continuous representations, we also devise a simple schema generation method which produces better schemas compared to a prior discrete representation based method. Our analysis shows that the tensors capture distinct usages of a predicate even when there are only subtle differences in their surface realizations.
\end{abstract}

\section{Introduction}
\noindent 
Understanding the events described in text is central to applications in artificial intelligence such as question answering, discourse understanding, and information extraction. Research in event understanding ranges from relation extraction of individual events to full document understanding of all its events.
Inspired by the concept of 
\textit{scripts}, proposed in the seminal work by \citet{schank77}, much work has looked at modeling stereotypical sequences of events in order to drive discourse understanding.
Early rule-based methods for this task were characteristically brittle and domain-specific.
Later work proposed computational models for script learning and understanding~\cite{mooney,chambers_schemas,balasubramanian2013generating}, but they use shallow event representations dependent on their specific surface words.
Others have focused on training neural networks for robust event representations, using them to predict which events are likely to occur next~\cite{modi16,pichotta16}.


To be broadly useful, a good representation should capture both event-level semantics (e.g., synonymy) and broader scenario-level semantics. Event-level semantics are accessible to simple composition methods. However, modeling scenario-level semantics is challenging. For example, consider the following events: (i) \emph{she threw a football} and (ii) \emph{she threw a bomb}. Even though the subject and the verb are the same, these two events denote entirely different scenarios -- sports and terrorism. The interaction of \emph{football} or \emph{bomb} with the predicate \emph{threw} is what determines the precise semantics of the event and the broader scenario in which it is embedded. A change to a single argument requires a large shift in the event's representation.
\begin{figure}[t!]
    \centering
    \includegraphics[scale=0.2]{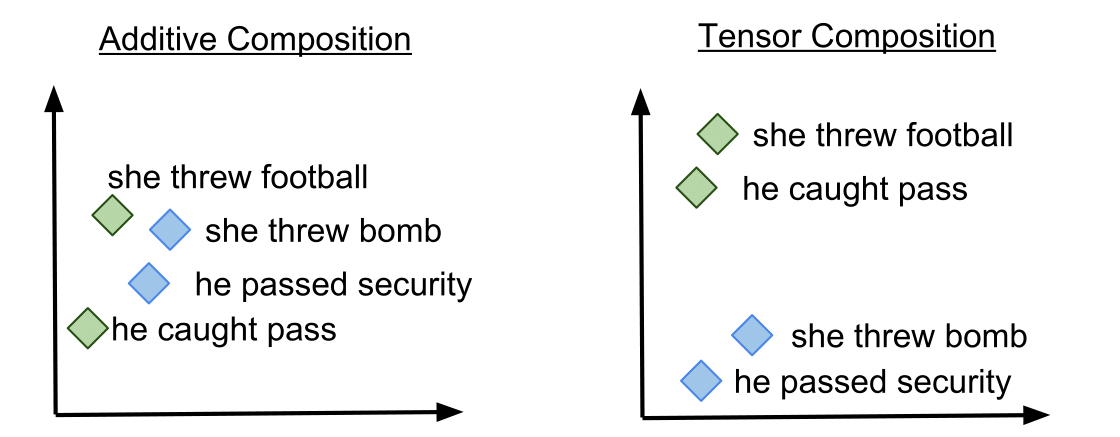}
    \caption{\label{fig:motivation} Additive compositions of word embeddings do not distinguish distinct events. Tensor compositions can tease out distinct events even with high lexical overlap, and recognize related events even with low lexical overlap.}
\end{figure}

As illustrated in Figure~\ref{fig:motivation}, due to the overlap of words, parameterized additive models~\cite{Granroth-Wilding16,modi16} and RNN-based models~\cite{pichotta16,hu_lstm_script} are limited in their transformations. Additive models combine the words in these phrases by the passing the concatenation or addition of their word embeddings to a parameterized function (usually a feed forward neural network) that maps the summed vector into event embedding space. The additive nature of these models makes it difficult to model subtle differences in an event's surface form.
Instead of additive models, we propose tensor-based composition models, which combine the subject, predicate, and object to produce the final event representation. The models capture multiplicative interactions between these elements and are thus able to make large shifts in event semantics with only small changes to the arguments. 

This paper puts forth three main contributions:
\begin{itemize}[nolistsep,noitemsep]
\item A scalable tensor-based composition model for event representations, which can implicitly recognize different event contexts based on predicate argument interactions.

\item A broad set of evaluations on event-related tasks: we find that the tensor-based event representation outperforms additive compositions on (i) a sentence similarity task, (ii) a new \textit{hard} similarity task, and  (iii) an event prediction task (two variants of the narrative cloze), suggesting broad utility in event-related tasks.

\item A simple yet effective method for generating event schemas: to our knowledge, ours is the first proposal for using continuous event representations in schema generation. 
We show that event tensors produce superior schemas compared to a prior distributional counting model~\cite{balasubramanian2013generating}. 

\eat{\item New event datasets: We release our initial dataset of event scripts generated from tensor-based representations and a small dataset of "hard event similarity" pairs.}

\eat{design a new "hard event similarity" 
release our new dataset of "hard event similarity" pairs, in addition to our initial dataset of event scripts generated from tensor-based representations.}
\end{itemize}

\section{Models: Tensor-based Event Composition}

Event composition models produce vector representations of an event given its predicate and argument representations. To be broadly useful, a representation should have two properties: (i) Events that are usually part of the same underlying scenario should get similar representations -- that is they should be embedded close together in the event space. (ii) Events that occur in different scenarios should get different representations, even if they have similar lexical expressions. In our earlier example, we want \emph{she threw football} and \emph{she threw bomb} to be farther apart in the embedding space, even though they share the subject and the verb.

An effective event composition model must be able to account for the numerous usage contexts for a given predicate and should be able to invoke the appropriate usage context based on the arguments. This problem is compounded even further when considering predicates that take on many different meanings depending on their arguments. This requires that the composition models be sensitive to small changes in the predicate argument interactions. Simple averaging or additive transformations are not enough. 

One way to capture this rich interaction is through tensor composition. Tensor-based composition has been used for composing over tree based structures ~\cite{socher2013recursive} and in general for modeling words with operator like semantics~\cite{Grefenstette2013MultiStepRL,fried2015low}. A key benefit of tensor composition that is relevant to our setting is that they capture multiplicative interactions of all the elements involved in the composition, which allows for the composition to be sensitive to even small changes in predicate argument interaction. 

Formally, we can define a tensor-based event composition model as follows. Given a predicate-specific tensor $P$, and the embeddings of the subject $s$ and the object $o$, the representation for the event $e$ can be composed through tensor contraction denoted as $e = P(s, o)$. Each element in the $d'$-dimensional event vector is obtained by the multiplicative interaction of all elements of the subject and object vectors, weighted by a term that depends on the predicate. The $ith$ component of  $e$ is computed as:
\begin{equation}
e_i = \sum_{j,k} P_{ijk} s_j o_k    
\end{equation}

The key question with predicate-specific tensors is how to learn and reason with such a huge number of parameters. Prior work that models words as tensors are not scalable because
the methods for learning such word-specific tensors require a large number of parameters, as well as a large number of training instances for each word. Relaxing the need for predicate specific training data and reducing the number of parameters motivate the two models we propose here.


\subsection{Predicate Tensor Model}
\begin{figure}[t!]
    \centering
    \includegraphics[scale=0.19]{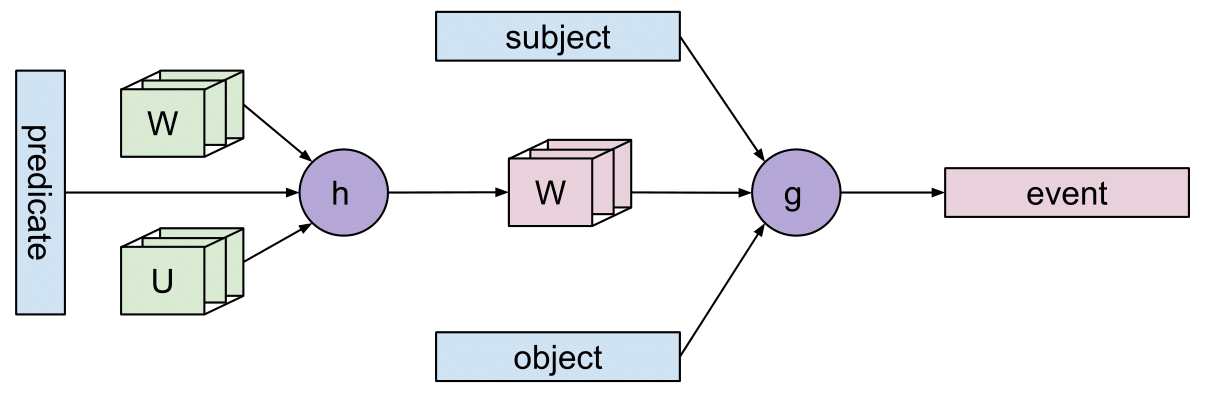}
    \caption{\label{fig:event-tensor}\small Predicate Tensor Model: Event representations are computed using a predicate tensor that combines the arguments. Elements in blue are inputs, green are model parameters, and pink are outputs generated by the model. Function $h$ produces the predicate tensor using tensors W and U as shown in Equation~\ref{eqn:predicate-tensor}. Function $g$ produces the final event representation through tensor contraction as shown in Equation~\ref{eqn:pt-contraction}.}
\end{figure}
Rather than learn a predicate-specific tensor, we instead learn two general tensors that can \emph{generate} a predicate-specific tensor on the fly. 
The predicate tensor model defines a function from a basic word embedding for the predicate to a tensor $P$. In this model, the predicate tensors are derived from a shared base tensor $W \in \mathbb{R}^{d \times d \times d}$ (where $d$ is the input embedding dimension). To allow the predicate's word embedding to influence its resulting tensor, we allow each element of $W$ (each one dimensional `row' of $W$) to be scaled
by a value that depends on a linear function of the predicate embedding $p$: 
\begin{equation}
    P_{ijk} = W_{ijk} \sum_{a} p_a U_{ajk} 
    \label{eqn:predicate-tensor}
\end{equation}

Here $U$ is also a tensor in $\mathbb{R}^{d \times d \times d}$ which defines linear functions for each one dimensional row of $W$, determining how 
$p$ should scale that dimension.
Now, given the predicate vector $p$, the subject vector $s$, and the object vector $o$, the original tensor contraction $P(s,o)$ we seek above is realized as follows.
Each element in the resulting event vector $e_i$ is computed as:

\begin{equation}
e_i = \sum_{a,i,j,k} p_a s_j o_k W_{ijk}U_{ajk} 
\label{eqn:pt-contraction}
\end{equation}

Thus this model captures multiplicative interactions across all three: subject, verb, and object.

\subsection{Role Factored Tensor Model}

\begin{figure}[t!]
    \centering
        \includegraphics[scale=0.15]{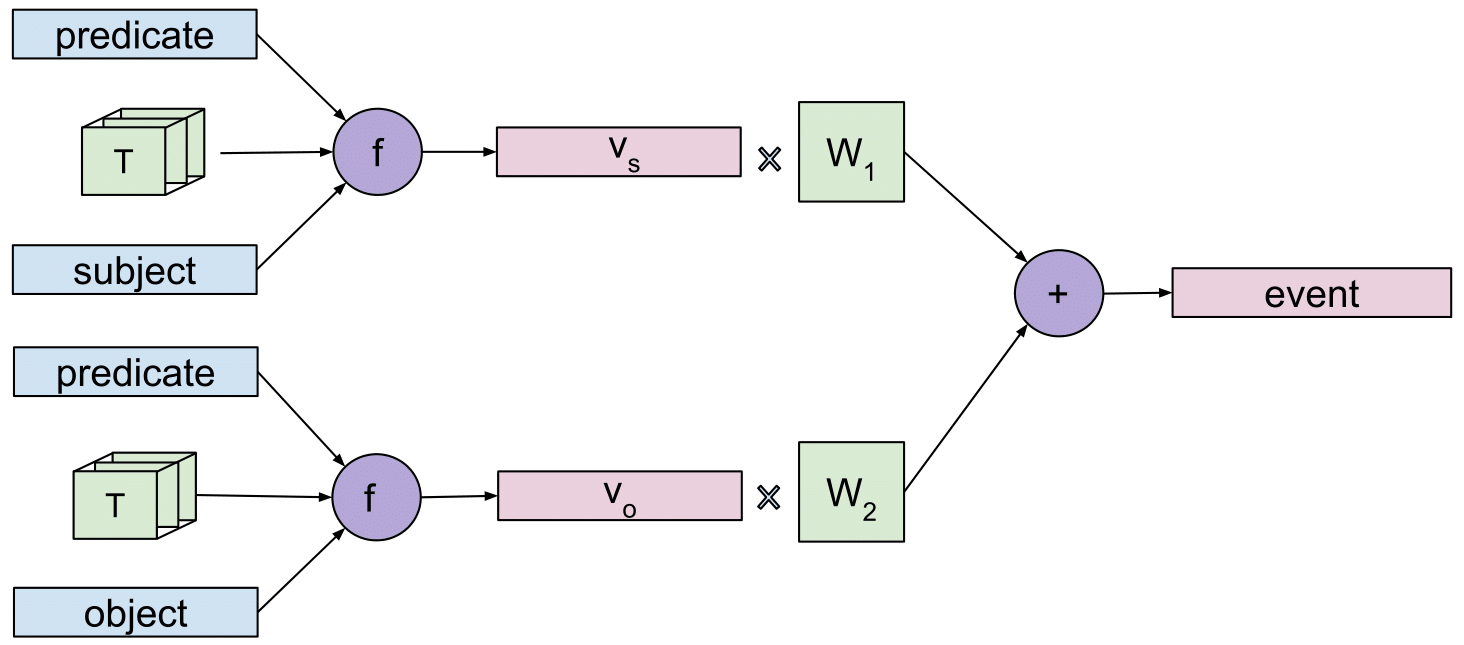}
    \caption{\small \label{fig:factored-event-tensor}Role Factored Tensor Model: Uses two tensor contractions of the predicate with the subject and object, which are then combined. Elements in blue are inputs, green are model parameters, and pink are outputs generated by the model. Function $f$ is the tensor contraction operation defined in Equation~\ref{eqn:ft-contraction}.}
\end{figure}
The representation power of the predicate tensor comes at the cost of model complexity and sensitivity. The ability to capture multiplicative interactions across all its arguments allows the predicate tensor to model complex relations between the predicate and its arguments. While this is useful from a representation stand point, it also makes the model behavior more complex and highly sensitive -- changes in the components of the event can drastically change the value of the event embedding.

In many cases, however, it may not be necessary to jointly model the interaction of a predicate with {\em all} of its arguments. Often times, the predicate's interaction with just one of its arguments is enough for recognizing its usage context. For example, knowing that \textit{football} is an  argument to \textit{throw} immediately places the event in the context of sports. 

Rather than model interactions between all components, we thus introduce a factored composition strategy: capture interactions between the predicate and its arguments separately to then combine these interactions into the final embedding. In this model, a single tensor $T \in \mathbb{R}^{h \times d \times d}$ (where $h$ is the output dimension and $d$ is defined as before) is used to capture these interactions. The resulting interactions are then combined using role-specific transformations\footnote{This approach is similar to the application of recursive tensor composition~\cite{socher2013recursive} but is not fully recursive in that the partial interactions are not combined using recursive tensor application. We found this additional recursion to be detrimental to performance.\nb{Does this read ok?}}. 

Formally, let $s$, $o$, and $p$ denote the word vectors for the subject, object, and the predicate respectively. The argument specific interactions are captured using two compositions of the predicate, one with the subject and one with the object: 
\begin{align}
v_s &= T(s,p) \\ 
v_o &= T(o,p)
\end{align}
As before, this composition is defined via tensor contraction. Given arbitrary arguments $a$ and $b$, the element $v_i$ of the output vector $T(a,b)=v \in \mathbb{R}^h$ is given by:
\begin{equation}
v_i = \sum_{j,k} T_{ljk} a_j b_k   
\label{eqn:ft-contraction}
\end{equation}

The subject and object interactions $v_s$ and $v_o$ are then  transformed via their respective role specific matrices, $W_s,W_v \in \mathbb{R}^{d'\times h}$ and summed together to obtain the final event embedding $e \in \mathbb{R}^{d'}$:
\begin{equation}
    e=W_s v_s + W_o v_o 
\end{equation}
Figure~\ref{fig:factored-event-tensor} displays the structure of the model. This factored model has fewer parameters compared to the Predicate Tensor Model and can also readily generalize to higher-arity events with any arbitrary number of arguments.

\section{Training Tasks}

In order to learn the parameters of the tensor composition models we employ two different training tasks, which correspond to predicting two different types of contexts. 

\subsection{Predict Events}
One natural way to learn script knowledge is to directly learn how to predict what other events are likely to occur given a set of events~\cite{pichotta16,modi16}. We use an alternate learning formulation defined over a pair of events rather than over sequences of events. One goal is to learn representations that maximize some similarity measure between co-occurring events similar to the task defined by~\citet{Granroth-Wilding16}\footnote{The objective function in \cite{Granroth-Wilding16} differs from ours in that they learn a coherence function which is used to measure similarity between events, rather than directly using cosine similarity.}. 

Given some input event $e_i$, a target event $e_t$ which occurs within in a window of $w$ in front of $e_i$ is randomly chosen. 
A negative event $e_n$ is also randomly sampled from the entire corpus. The regularized learning objective is to minimize the following quantity:
\[
    \frac{1}{N} \sum^{N}_{i=1} max(0,m + sim(e_i, e_n) - sim(e_i, e_t)) + \lambda L(\theta)
\]
where $sim(a,b)$ is cosine similarity, $m$ is the margin, and $L(\theta)$ is $l2$ regularization on all model parameters $\theta$.

\subsection{Predict Words}
A second training approach is to predict the nearby words of an event's sentence context, rather than just the event words. This is similar to most other word representation learning objectives.
Formally the objective function to minimize is:
\[
 \frac{1}{N} \sum^N_{i=1} \sum^{n_{e_i}}_{j=1} -log(P(w_{ij} | e_i)) + \lambda L(\theta)
\]
where $n_{e_i}$ is the number of words in the sentence that $e_i$ appears in, and $w_{ij}$ is the $j^{th}$ word in 
the sentence for $e_i$. $P(w_{ij} | e_i)$ is computed via a softmax layer.
The use of this type of inter sentence word prediction for finding embeddings for phrases has been shown to be useful \cite{paragraphVects}.

\subsection{Training Details}
We use the New York Times Gigaword Corpus for training data. Event triples are extracted using the Open Information Extraction system Ollie \cite{schmitz2012open}. We initialize the word embedding layer
with 100 dimensional pretrained GloVe vectors \cite{pennington2014glove}. Like previous work \cite{modi16,Granroth-Wilding16}, we allow the word embeddings to be further updated during training for all models.

We hold out 4000 articles from the corpus to construct dev
sets for hyperparameter tuning, and 6000 articles for test purposes. We make all code and data publicly available.\footnote{\texttt{github.com/stonybrooknlp/event-tensors}}
Hyperparameter tuning was done using dev sets constructed for the CMCNC and MCNC tasks (see below). Training was done using Adagrad \cite{adagrad} with a learning rate of 0.01  
and a minibatch size of 128. 


\section{Evaluation}
\noindent We evaluate our proposed tensor models on a variety of event related tasks, comparing against a compositional neural network model, a simple multiplicative model, and an averaging baseline. Since our 
goal is to produce a event representation given a single event, and not a sequence as in \citet{pichotta16}, RNN based models are not suitable for this task.
We report results on both training tasks: predicting events (EV) and predicting words (WP).

\subsection{Baselines}
{\bf Compositional Neural Network (NN)} Our first baseline is an neural network model used previously in recent work. 
The event representation in this model is computed by feeding the concatenation of the subject, predicate, and object embedding into a two layer neural network\footnote{We additionally tried a three layer network, however this slightly decreased performance on the dev set}. 
\[
e = W*tanh(H[s;p;o])
\]
where $W$ and $H$ are matrices, the main model parameters.

This basic architecture was used to generate event representations for narrative cloze tasks~\cite{modi13,modi16,Granroth-Wilding16}. We adapt this architecture for our task -- given the input event, whose representation is to be composed, predict the neighboring context (other events or words in the sentence). This deviates from the model used in \cite{modi16} mainly in training objective, and that the embedding of the protagonist's dependency is fed in as input (see the Related Works section for details). This architecture was also used in \cite{Granroth-Wilding16} to obtain event representations, which are then fed to another network that measures coherence of the event representations. We only use the event representation part of this model and train it to maximize our training objectives.

\noindent{\bf Elementwise Multiplicative Composition} The second baseline extends the additive
composition model by simply concatenating the elementwise multiplications between the verb and its subject/object. This models some (though not all) of the multiplicative interactions. The embedding in this model is computed as:
\[
e = W*tanh(H[s;p;o;p\odot s;p\odot o])
\]

\noindent
where $\odot$ denotes elementwise multiplication.

\noindent{\bf Averaging Baseline} This represents each event as the average of the constituent word vectors using pretrained GloVe embeddings \cite{pennington2014glove}. 

\subsection{Similarity Evaluations}

\subsubsection{Transitive Sentence Similarity}

\begin{table}[t!]
\centering
\begin{tabular}{|c|c|c|} 
 \hline
 \multirow{2}{*}{System}& \multicolumn{2}{|c|}{$\rho$} \\
 \cline{2-3}
 & WP & EV \\
 \hline
 Role Factor Tensor & \textbf{0.71} & \textbf{0.64}\\ 
 Predicate Tensor & \textbf{0.71} & 0.63 \\
 Comp. Neural Net & 0.68 & 0.63 \\
 Elementwise Multiplicative & 0.65 & 0.57 \\
 \hline
 Averaging & \multicolumn{2}{|c|}{0.67}\\
 \hline
\end{tabular}
\caption{Performance on the Transitive Sentence Similarity dataset, as indicated by Spearman's $\rho$}
\label{table:trans}
\end{table}

Similarity tasks are a common way to measure the quality of vector embeddings. 
The transitive sentence similarity dataset \cite{transitive} contains 108 pairs of transitive sentences: short phrases containing a single subject, object, and verb (e.g., \textit{agent sell property}). Every pair is annotated by a human with a similarity score from 1 to 7. For example, pairs such as 
\textit{(design, reduce, amount)} and \textit{(company, cut, cost)} are annotated with a high similarity score, while pairs such as \textit{(wife, pour, tea)} and \textit{(worker, join, party)} are given low 
similarity scores. Since each pair has several annotations, we use the average annotator score as the gold score. We evaluate using the Spearman's correlation of the cosine similarity given by each model and the annotated similarity score.

\subsubsection{Hard Similarity Task}

The main requirement we laid out for event composition is that similar events should be embedded close to each other, and dissimilar events or those from distinct scenarios should be farther from each other. We create a hard similarity task to explicitly measure how well the representations satisfy this requirement. To this end, we create two types of event pairs, one with events that should be close to each other but have very little lexical overlap (\textit{e.g., police catch robber / authorities apprehend suspect}), and another with events that should be farther apart but have high overlap  (\textit{e.g., police catch robber / police catch disease}. 

A good composition model should have higher cosine similarity for the similar pair than for the dissimilar pair. To evaluate this directly we created 230 pairs (115 pairs each of similar and dissimilar types).
To create the set, we have one annotator create similar/dissimilar pairs from a set of ambiguous verbs, while three different annotators give the similarity/dissimilarity rankings. We kept pairs where the annotators agreed completely.
For each composition method, we obtain the cosine similarity of the pairs under that representation, and report the fraction of cases where the similar pair receives a higher cosine than the dissimilar pair.

\eat{We have given examples of events that are difficult for compositional 
models to differentiate (\textit{e.g.} \textit{chef cooked books} vs \textit{chef
cooked meal}). In order to test a model's ability to do so, we propose a new task
called the \textit{hard similarity task}. Given two pairs of 
events, a system must decide what pair is most similar using only the cosine distance
between their representations. Like the cooking example in the Introduction, a pair of events might differ in a single word, yet are highly different (\textit{e.g., police catch robber / police catch disease}). The other pair are highly similar, but differ in word choice (\textit{e.g., police catch robber / authorities apprehend suspect})
The task is treated as a classification task, where the correct class is the 
semantically similar but lexically different pair. The dataset is composed of 230
pairs (115 instances to classify). We will release this dataset upon paper acceptance.}

\eat{\begin{table}[t!]
\centering
\begin{tabular}{|c|c|c|} 
 \hline
 \multirow{2}{*}{System}& \multicolumn{2}{|c|}{$\rho$} \\
 \cline{2-3}
 & WP & EV \\
 \hline
 Role Factor Tensor & \textbf{0.71} & \textbf{0.64}\\ 
 Predicate Tensor & \textbf{0.71} & 0.63 \\
 Comp. Neural Net & 0.68 & 0.63 \\
 Elementwise Multiplicative & 0.65 & 0.57 \\
 \hline
 Averaging & \multicolumn{2}{|c|}{0.67}\\
 \hline
\end{tabular}
\caption{Performance on the Transitive Sentence Similarity dataset, as indicated by Spearman's $\rho$}
\label{table:trans}
\end{table}}

\begin{table}[t!]  
\centering
\begin{tabular}{ |c|c|c| } 
 \hline
 \multirow{2}{*}{System}& \multicolumn{2}{|c|}{Accuracy} \\
 \cline{2-3}
 & WP & EV \\
 \hline
 Role Factor Tensor & 23.5 & \textbf{43.5}$\dagger$ \\
 Predicate Tensor & \textbf{35.7}$\dagger$ & 41.0 $\dagger$ \\
 Comp. Neural Net & 20.9 & 33.0 \\ 
 Elementwise Mult. & 11.3 & 33.9 \\
 \hline
 Averaging & \multicolumn{2}{|c|}{5.2}\\
 \hline
\end{tabular}
\caption{Hard Similarity Results: Accuracy is the percentage of cases where the similar pair had higher cosine similarity than the dissimilar pair.}
\label{table:hard_sim}
\end{table}

\subsubsection{Results}

Table~\ref{table:trans} shows the Spearman's $\rho$ scores of the various models on the transitive similarity task.
Consistent with prior work~\cite{milajevs}, we find that simple averaging is a competitive baseline for the task. The neural network baseline is only slightly better than averaging. When training for word prediction, both the predicate and the role factored tensor models have a higher correlation (+3 points in $\rho$). Across all models, training for the word objective is better than training for the event objective. This is to be expected since the word prediction task focuses on a narrower context,
requiring that representation to be consistent with information within the sentence in which it appears. This goal aligns well with the sentence similarity task where the events that are roughly exchangeable are deemed similar.

Table~\ref{table:hard_sim} compares the percentage of cases where the similar pairs had a higher cosine similarity than dissimilar pairs under each event composition method. Both tensor methods outperform the baselines showing that tensors capture a richer and broader set of semantics about the events. Interestingly, the event objective is better than word prediction for all models. We believe the broader context used by the event prediction objective helps capture broader event semantics necessary to generalize to the difficult pairs.


\subsection{Coherent Multiple Choice Narrative Cloze}
The above experiments judge event similarity, but our broader goal is to model real-world knowledge.
The \emph{narrative cloze task} was proposed to evaluate script knowledge and knowledge about events that occur together in the world \cite{chambers_chains}. 
The task starts with a series of events that are mentioned in a document, but hides one of the events.
A reasoning system should predict what the held out event is, given only the other events. 
Event predication is typically done by choosing an event such that it maximizes some similarity score 
with the context events.\footnote{We use cosine similarity as our measure}
This formulation is difficult since the output space of possible events is rather large.  
We evaluate on variant called the multi-choice narrative cloze (MCNC)~\cite{Granroth-Wilding16}, where the system should distinguish the held out event from a small set of randomly drawn events. 


However, using automatic cloze evaluation has multiple issues~\cite{badcloze}, one of which is that it is overly sensitive to frequency cutoffs of common events (e.g. \textit{said}, \textit{was}, \textit{did}, etc.) and errors in the preprocessing tools. 
To address these, we manually curated an evaluation set for the MCNC task, and call it the Coherent Multiple Choice Narrative Cloze (CMCNC). 
\begin{table}[t!]
\centering
\begin{tabular}{ |c|c|c|c|c| } 
 \hline
 \multirow{2}{*}{System}& \multicolumn{2}{|c|}{CMCNC} & \multicolumn{2}{|c|}{MCNC} \\
 \cline{2-5}
 & WP & EV & WP & EV\\
 \hline
 Role Factor Tensor & \textbf{70.1}$\dagger$ & \textbf{72.1} $\dagger$ & \textbf{42.2} & \textbf{46.5} $\dagger$\\
 Predicate Tensor & 64.5 & 66.1 & 32.3 & 41.1\\
 Comp. Neural Net & 65.7 & 68.5 & 38.4 & 45.3\\ 
 Elementwise Mult. & 67.3 & 67.7 & 41.7 & 45.4 \\
 \hline
 Averaging & \multicolumn{2}{|c|}{26.7} & \multicolumn{2}{|c|}{14.3}\\
 \hline
\end{tabular}
\caption{CMCNC Results: Predict the held out event, given the observed events from the same document. The dataset is curated manually for coherence by removing noisy instances and context events. $\dagger$ denotes a statistically significant difference($\alpha < 0.05$) over the best competing baseline under a paired t-test.}
\label{table:cmcnc}
\end{table}
We generated an MCNC dataset and made three modifications.

First, we manually removed events that are either: (1) frequent events types that are in our stop event list \nb{(e.g., )}, or (2) non-sensical events coming from obvious extraction errors \nb{(e.g., )}.

Second, we discard heldout events that don't fit the following criteria: (i) One of its entities must appear in the given context events. (ii) Given the context events, the held out event should seem plausible to a human evaluator. 

\noindent Third, when the heldout event is included with 5 randomly selected negative events, we replace the entities appearing in the negative events with similar (under word embedding distance) entities that appear in the context events. 

Enforcing these constraints allows us to better evaluate how well the system learned commonsense script
knowledge, rather than how well the system learned to emulate noise in the data. For comparison, we also report on an automatically generated MCNC task. Since we do not restrict 
the evaluation to narrative chains with a single protagonist, the numbers for the 
automatic MCNC task are lower than those reported in \citet{Granroth-Wilding16}.

\subsubsection{Results}

\eat{
\begin{table}[t!]
\centering
\begin{tabular}{ |c|c|c| } 
 \hline
  \multirow{2}{*}{System}& \multicolumn{2}{|c|}{Accuracy} \\
 \cline{2-3}
 & WP & EV \\
 \hline
 Role Factor Tensor & \textbf{42.2} & \textbf{46.5} \\ 
 Comp. Neural Net & 38.4 & 45.3 \\ 
 Predicate Tensor & 32.3 & 41.1 \\
 Elementwise Multiplicative & 41.7 & 45.4 \\
 \hline
  Averaging & \multicolumn{2}{|c|}{14.3}\\
 \hline
\end{tabular}
\caption{Automatic MCNC Results: Task is to predict a held out event, given a set of context events from the same document. $\dagger$ indicates statistically significant differences over Compositional Neural Network model.}
\label{table:mcnc}
\end{table}}

Table~\ref{table:cmcnc} show the results for the manually filtered CMCNC and the automatic MCNC tasks. 
Whereas the word-based objective excelled on similarity tasks, the event-based objective does better on cloze tasks. The Role Factor model shows significant improvements in accuracy compared to the neural network model (+4.4 points with word prediction and +3.6 points with event prediction). The Predicate tensor model, however, performs worse than the neural network model.
The human-curated evaluation shows vastly different results than the automatic-MCNC, perhaps bolstering the idea that MCNC is not ideal~\cite{badcloze}. 
Our Role Factor model excels in both. 

\section{Generating Event Schemas}

An event schema is a form of script like knowledge about a scenario (e.g., a \textit{bank heist}). \citet{chambers_schemas} introduce schemas as a set of the main entities in the scenario (the \textit{robber, bank teller, etc.}) and the main events they participate in (\textit{robber steals money, teller calls police etc.}).

Prior work on event schema generation use discrete representations of events and build count-based conditional models over these representations~c.f., \cite{chambers_schemas,balasubramanian2013generating}. The basic idea behind these methods is to start with a seed event and locate other events that are also likely to occur in the scenario. These methods suffer fragmented counts because of synonymy (different tuples can denote the same event) and mixing events from different contexts because of polysemy (similar looking tuples denote distinct events). Continuous event representation provides an opportunity to address these issues. 

The tensor-based representations yields a simpler and more direct approach to generating event schemas. Recall that with the predict events objective, events that tend to co-occur with each other end up with a similar representation and get embedded closer to each other in the event space. To create a schema seeded by an event, we can simply find its nearest neighbors in the event embedding space and use them as candidates. This method is simple, scalable, and does not require maintaining large tables of co-occurrence information (c.f.~\citet{balasubramanian2013generating}).


\eat{An event schema is a structured knowledge source that define a scenario (for example, a \textit{Bank Heist}) 
by describing the entities that participate in the scenario and the typical roles they play (the \textit{robber, bank teller, etc.}), the relationships between these entities, and the actions they take during the scenario (\textit{robber steals money, bank teller gives money}). Event schemas are helpful in tasks such as information extraction, where they can help inform the system about the type
of information it needs to look for. 

\cite{chambers_schemas} show that these schemas can be automatically generated by utilizing script knowledge (in their case, induced
via event co-occurrence counts). In this work, we have induced script knowledge via continuous vector representations.
One natural question to ask is whether event schemas can be automatically generated using these representations, and if so, do 
they perform better then those based on count based approaches?
In this section we provide a simple way to generate schemas via our event representations using nearest neighbor search. Despite 
its simplicity, we find that the schemas it produces are of a much higher quality when compared to those produced by a count based 
approach. 

}

\subsubsection{Nearest Neighbor Schema Generation}
Like previous systems, our system takes as input a single seed event and produces the schema based on this event. Given a seed
event $s$ the algorithm proceeds as follows:
\begin{itemize}[nolistsep,noitemsep]
    \item From a corpus of events, compute their representations using one of the composition methods described above
    \item Find the $k$ nearest neighbors to $s$ in the text corpus with respect to the cosine distance of their representations.
    \item Go through this list of nearest neighbors, add a neighbor $x$ to the schema if all the following conditions are true:
    \begin{itemize}
        \item The cosine distance between the GloVe embedding of $x$'s predicate and all other predicates currently in the schema is greater than $\alpha$
        \item The cosine distance between the GloVe embedding of at least one of $x$'s arguments and some entity $e$ in the schema is less than $\beta$. The argument is then replaced with $e$ to create a new event $x'$
        \item The average cosine distance between the representation of $x'$ (computed using the same composition function used to compute the original representations) with all other events in the schema is less
        than $\gamma$
    \end{itemize}
\end{itemize}

\begin{figure}[t!]
    \centering
    \includegraphics[scale=0.19]{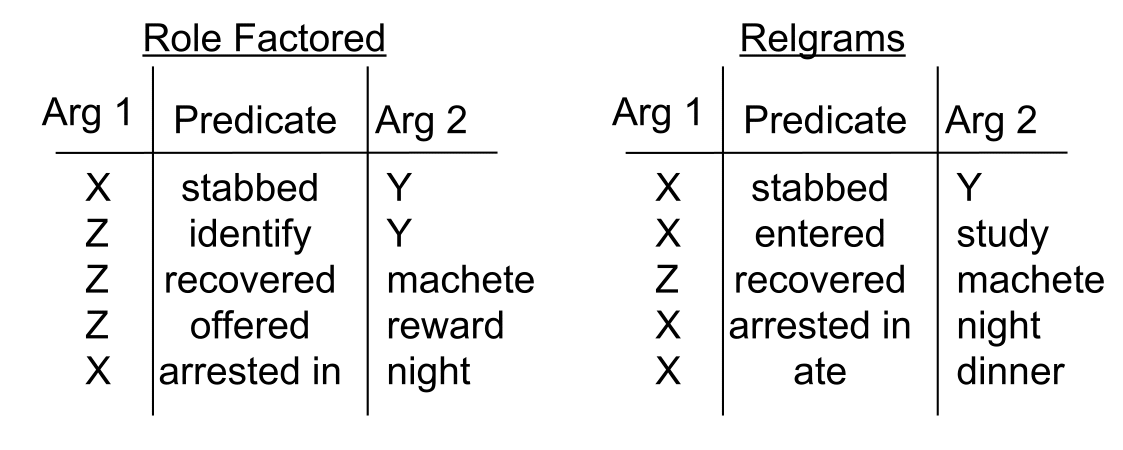}
    \caption{\label{fig:schemas}Example schemas from nearest neighbors and relgrams approach. X, Y, and Z are variables that denote distinct entities. Each row indicates the main event that the entities participate in.}
\end{figure}

\begin{table}
\centering
\begin{tabular}{ |c|c|c| } 
 \hline
 \multirow{2}{*}{System}& \multicolumn{2}{|c|}{Average Score} \\
 \cline{2-3}
 & With 0's & Without 0's\\ 
 \hline
 Role Factor (EV) & \textbf{2.45} & \textbf{2.62} \\
 Comp. Neural Net (EV) & 2.26 & 2.47 \\
 \cite{balasubramanian2013generating} & 1.51 & 1.78 \\
\hline
\end{tabular}
\caption{Average Annotator Scores for generated schemas (scale from 0-4). Higher scores indicate a more coherent schema.}
\label{table:schema_eval}
\end{table}

The end result is a series of events representing a schema. For values of $\alpha, \beta$ and $\gamma$, we use $\alpha=0.5$, 
$\beta=0.25$, $\gamma=0.2$ (for Role Factor), and $\gamma=0.3$ (for Neural Network). Values were selected using a develoment 
set of 20 seeds.

\subsection{Schema Evaluation}
We compare against previous work that produces schemas using relation co-occurrence graphs~\cite{balasubramanian2013generating}. It also uses OpenIE triples mined from a text corpus. We refer to this as the \textit{Relgrams approach}.
For a fair comparison, we choose 20 seed tuples at random from the list of top seeds in~\cite{balasubramanian2013generating}. For each seed, we generate a schema of 10 grounded events using our best models. We report results using the Role Factor (EV) representations, the Compositional Neural Network (EV) representations, and the Relgrams approach.

Our evaluation is human-driven. We present annotators with 3 schemas (from the 3 models) using the same seed. 
For each event in each schema, annotators rated the event on a 0-4 scale with regards to its relevance to the seed and the rest of the schema. A score of 0 is reserved for events that are completely nonsensical (either caused by an extraction error, or bad entity replacement). A score of 1 indicates that the event was not relevant with no obvious relation to the scenario, while a score of 4 indicates the event is highly relevant and would be a core part of any description of the scenario. Table \ref{table:schema_eval} shows the average scores given for each method. 
We report average ratings both with and without nonsensical events\footnote{All differences in means are significant under t-test w/ $\alpha<.05$}.

Both nearest neighbor schemas consistently out-ranked the Relgrams schemas by a large margin. Figure~\ref{fig:schemas} shows the schemas produced by both systems using the same seed. The biggest problem with the schemas produced by the Relgrams method was their tendency to include overly specific triples in the schemas, simply because they co-occur with one of the events in the schema. The schemas based on continuous representations can avoid this problem. Here is a typical example: the event \textit{(police, found, machete)} co-occurs a couple times
in an article with \textit{(he, likes to eat, dinner)}. Since \textit{(police, found, machete)} is rare in the corpus, the count based system takes into consideration the co-occurrence. Although tuples similar to \textit{(police, found, machete)} such as \textit{(authorities, recovered, murder weapon)} may appear in the corpus, the counts based on discrete representations cannot share evidence across these similar tuples. The method based on continuous representations can take advantage of similarity of words in the two tuples. 
The tensor model further aids by helping the system better differentiate between events with many similar words, but different meanings. 

While the Compositional Neural Network and Role Factor representations both perform well compared to the Relgrams approach, the Role Factor factor does better, suggesting that the model's ability to tease out contextual usages of predicates can be useful in end tasks. 

\eat{\begin{table}
\centering
\begin{tabular}{ |c|c|c| } 
 \hline
 \multirow{2}{*}{System}& \multicolumn{2}{|c|}{Average Score} \\
 \cline{2-3}
 & With 0's & Without 0's\\ 
 \hline
 Role Factor (EV) & \textbf{2.45} & \textbf{2.62} \\
 Comp. Neural Net (EV) & 2.26 & 2.47 \\
 \cite{balasubramanian2013generating} & 1.51 & 1.78 \\
\hline
\end{tabular}
\caption{Average Annotator Scores for generated schemas (scale from 0-4). Higher scores indicate a more coherent schema.}
\label{table:schema_eval}
\end{table}}

\section{Discussion}
This is the first paper to propose and evaluate event models on both event similarity and event schema prediction/generation.
We believe there is significant overlap in goals (any event representation needs to learn a nearness between similar events), but there are also differences (script learning is not at all about synonymy).

\subsection{Interpreting the Role Factor Model}
\eat{\nb{Add a line summarizing all the results. Attribute it to the main strength of the role factored model, which is the ability to capture different usage contexts of a predicate. Here we illustrate how the role factored model handles multiple usage contexts.}}
The Role Factor model consistently outperforms on all tasks, in particular over additive/concatenation models from recent work in event learning.
One question to ask is what exactly is the Role Factor model learning? 

Besides simply allowing multiplicative interactions between the predicate and its arguments, the Role Factored model can be interpreted as capturing different scenarios or contexts in which a predicate is used.
First, observe that performing the contraction to compute $v=T(a,p)$ can be done by first partially applying $T$ to $p$, producing a matrix $P$, whose columns $P_i$ are the result of multiplying each slice $T_i$ in $T$ by the predicate vector $p$. Thus, each slice $T_i$ moves the predicate vector to a new point $p'$ in the original word embedding space. If there are distinct usages for a predicate $p$, we hope that the slices of $T$ provide a map from $p$ to those usages in word embedding space. 



Table~\ref{table:neighbors} shows the nearest neighbors for several different $P_i$ vectors for the verb \textit{throw}. Each of these nearest neighbor clusters indicate what type of arguments (either subject or object) should 'activate' the $i^{th}$ dimension, reflecting the type of event contexts that tensor $T$ captures. 

\subsection{Script Generation}
This is the first attempt to generate event scripts (or schemas) from continuous representations. 
Recent work focuses on event language models that can perform narrative cloze, but to our knowledge, the goal of the learning explicit scripts has been left behind.
Our human evaluation of generated scripts shows that nearest neighbor selection with coherent entities builds far superior scripts than previous work. Figure~\ref{fig:schemas} gives an example of a generated schema from both systems.

\begin{table}[t!]
\centering
\begin{small}
\begin{tabular}{|c|c|c|c|c|} 
 \hline
{\bf context 1} & {\bf context 2} & {\bf context 3} & {\bf context 4} & {\bf context 5}\\
\hline
 hitting & police & shouted & farve & lodged \\
 inning & marchers & chanted & elway & complaint\\
 walked & chechens & chanting & yards& filed \\
 hit & stoned & yelled & rookie & remand \\
 fielder & protesters & shouting & broncos & lawsuit \\
 \hline
\end{tabular}
\end{small}
\caption{Nearest Neighbors for the implicitly learned contexts in which 'throw' may be used.}
\label{table:neighbors}
\end{table}
\nc{Would be great to have a table with 2 full schemas, one good one bad.}

\section{Related Work}

{\bf Neural Event Representations} Neural event embedding approaches learn representations by training to predict the next event. \citet{Granroth-Wilding16} concatenate predicate and argument embeddings and feed them to a neural network to generate an event embedding. Event embeddings are further concatenated and fed through another neural network to predict the coherence between the events. \citet{modi16} encode a set of events in a similar way and use that to incrementally predict the next event -- first the argument, then the predicate and then next argument. 
\citet{pichotta16} treat event prediction as a sequence to sequence problem and use 
RNN based models conditioned on event sequences in order to predict the next event.
These three works all model \textit{narrative chains}, that is, event sequences in which a single entity (the \textit{protagonist}) participates in every event. 
\citet{hu_lstm_script} also apply a RNN approach, applying a new hierarchical LSTM model in order predict events by generating discriptive word sequences.

{\bf Multiplicative and Tensor composition models} Tensor based composition models have been shown to be useful for other NLP tasks such as sentiment analysis~\cite{Socher:2011:SRA:2145432.2145450,socher2013recursive}, knowledge base completion \cite{socher2013reasoning}, and in general for demonstrating compositional semantics in measuring sentence similarity~\cite{Grefenstette2013MultiStepRL,kartsaklis2014study,fried2015low,kim2015neural}.

\citet{polajnar2015exploration} learn verb specific tensors by setting up a regression task where the learned tensor for a verb is expected to yield a representation for the sentential contexts where the verb is found. 
\citet{liberalie} use tensor based autoencoders over AMR representations of events in order to induce event level (rather than 
scenerio level), ACE\footnote{\texttt{www.itl.nist.gov/iad/mig/tests/ace/}} style templates for 
use in event extraction. 
Similarly, \citet{huang2017zero} use tensor compositions to build representations that facilitate zero shot relation extraction. 
Related to our hard similarity task, \citet{tilk2016event} use (factorized) tensor based methods to model the thematic fit between event arguments.

\eat{
Low-rank tensors for verbs in compositional distributional semantics (2015) Fried et al. 
Same thing as above, but using factorized versions of the tensors
A Study of Entanglement in a Categorical Framework of Natural Language (2014)
Where the transitive sentence similarity data set comes from
An exploration of discourse-based sentence spaces for compositional distributional semantics (2015)
Show that there are potential benefits to using contextual words from surrounding sentences when building tensor based representations 
Neural word embeddings with multiplicative feature interactions for tensor-based compositions (2015)
Show that if using CBOW to get initial word embeddings, can get some improved results by multiplying context vectors (if using a multiplicative composition function)
}

{\bf Schema/script learning} Unsupervised learning of script knowledge can be traced back to \cite{chambers_chains}, which introduced
a count based technique for inducing narrative chains. \citet{chambers_schemas} extends this
idea to creating full on narrative schemas by merging together narrative chains with argument overlap. Other unsupervised induction approaches include a relation n-gram based method \cite{balasubramanian2013generating}, and generative latent variable models~\cite{nguyen2015generative,chambers2013event,cheung2013probabilistic}. All of these models worked on discrete representations for capturing co-occurrence statistics. In this work, we show that higher quality scripts can be produced using continuous representations instead.


\section{Conclusions}
Understanding events requires effective representations that contain both information that is specific to the event and information that relates to the underlying context in which the event occurs. We propose tensor-based composition models which are able to capture the distinct event contexts in which a predicate gets used. This improved modeling of these event contexts allows the resulting continuous representations to be more effective in multiple event related tasks. Last, we also show that the continuous representations yield a simple schema generation method which produces schemas superior to a prior count based schema generation method.
\subsection{Acknowledgements}
This work is supported in part by the National Science Foundation under Grant IIS-1617969. 


\eat{models that can recognize distinct usages of predicates in different event contexts.
The argument interactions with their predicate are crucial to our approach. The multiplicative interactions captured by the tensors enables the resulting representations to be more effective in event related tasks. Last, we also show that the continuous event representations yield a simple schema generation method whose schemas are superior to a prior count based schema generation method.}

\bibliographystyle{aaai}
\bibliography{tensors}
\end{document}